\documentclass[letterpaper, 10 pt, conference]{ieeeconf}  % Comment this line out if you need a4paper

\IEEEoverridecommandlockouts                              % This command is only needed if 
\usepackage{graphics} % for pdf, bitmapped graphics files
\usepackage{epsfig} % for postscript graphics files
\usepackage{mathptmx} % assumes new font selection scheme installed
\usepackage{times} % assumes new font selection scheme installed
\usepackage{amsmath} % assumes amsmath package installed
\usepackage{amssymb}  % assumes amsmath package installed
\usepackage{booktabs}   % For \toprule, \midrule, \bottomrule
\usepackage{colortbl}   % For \rowcolor
\usepackage{xcolor}  
\usepackage{caption}
\usepackage{cite}
\usepackage{graphicx}
\usepackage[ruled,vlined,linesnumbered,noend]{algorithm2e}
\usepackage{algpseudocode}
\usepackage{etoolbox}
\usepackage{bbding}
\DontPrintSemicolon
\SetKwInOut{Input}{Inputs}
\SetKwInOut{Output}{Output}
\SetKwComment{Comment}{$\triangleright$ }{}
\SetKwProg{Proc}{procedure}{}{}
\algrenewcommand\algorithmicrequire{\textbf{Inputs:}}
\algrenewcommand\algorithmicensure{\textbf{Output:}}
\algrenewcommand\algorithmiccomment[1]{\hfill\(\triangleright\)\,#1}

\title{\LARGE \bf
SEF-MAP: Subspace-Decomposed Expert Fusion for Robust Multimodal HD Map Prediction
}

\author{\textbf{Haoxiang Fu$^{1,*}$, Lingfeng Zhang$^{2,3,*}$, Hao Li$^{7,*}$, Ruibing Hu$^{4}$, Zhengrong Li$^{5}$} \\ 
\textbf{Guanjing Liu$^{6}$, Zimu Tan$^{7}$, Long Chen$^{3}$, Hangjun Ye$^{3}$, Xiaoshuai Hao$^{3, \text{\dag},\text{\ddag}}$}
\thanks{$^\ddag$ Project Leader.}
\thanks{$^\text{\dag}$ Corresponding Author.}
\\[4pt]
$^{1}$National University of Singapore
\\
$^{2}$Shenzhen Tsinghua International Graduate School, Tsinghua Univeristy
\\
\vspace{2pt}
$^{3}$Xiaomi EV, $^{4}$Chinese University of Hong Kong \\
\vspace{2pt}
$^{5}$The University of Manchester, $^{6}$Renmin University of China
\\
\vspace{2pt}
$^{7}$Independent Researcher
\vspace{2pt}
\\
{\tt\small haoxiaoshuai714@163.com}
\vspace{-13pt}
}

\begin{document}

\maketitle
\thispagestyle{empty}
\pagestyle{empty}

%%%%%%%%%%%%%%%%%%%%%%%%%%%%%%%%%%%%%%%%%%%%%%%%%%%%%%%%%%%%%%%%%%%%%%%%%%%%%%%%
\begin{abstract}
High-definition (HD) maps are essential for autonomous driving, yet multi-modal fusion often suffers from inconsistency between camera and LiDAR modalities, leading to performance degradation under low-light conditions, occlusions, or sparse point clouds. To address this, we propose \textit{SEF-MAP}, a Subspace-Expert Fusion framework for robust multi-modal HD map prediction. The key idea is to explicitly disentangle BEV features into four semantic subspaces: LiDAR-private, Image-private, Shared, and Interaction. Each subspace is assigned a dedicated expert, thereby preserving modality-specific cues while capturing cross-modal consensus. To adaptively combine expert outputs, we introduce an uncertainty-aware gating mechanism at the BEV-cell level, where unreliable experts are down-weighted based on predictive variance, complemented by a usage balance regularizer to prevent expert collapse. To enhance robustness in degraded conditions and promote role specialization, we further propose distribution-aware masking: during training, modality-drop scenarios are simulated using EMA-statistical surrogate features, and a specialization loss enforces distinct behaviors of private, shared, and interaction experts across complete and masked inputs. Experiments on nuScenes and Argoverse2 benchmarks demonstrate that \textit{SEF-MAP} achieves state-of-the-art performance, surpassing prior methods by +4.2\% and +4.8\% in mAP, respectively. \textit{SEF-MAP} provides a robust and effective solution for multi-modal HD map prediction under diverse and degraded conditions.
\end{abstract}

% \documentclass[twocolumn]{article}

% \begin{document}

\section{INTRODUCTION}

High-definition (HD) maps are crucial for autonomous driving, providing precise semantic and geometric information for perception, planning, and navigation~\cite{hao2025robust,elghazaly2023hdmap,li2022hdmapnet,wijaya2024hdmap, hao2024your, hao2025mapfusion, hao2025msc}. Unlike raw sensor data, HD maps offer structured road topology, lane geometry, and traffic elements, supporting reliable long-term autonomy~\cite{elghazaly2023hdmap,bao2022hdmap}. The rise of bird’s-eye-view (BEV) perception has enabled HD map construction directly from multi-view cameras and LiDAR. However, building robust and accurate maps remains challenging due to the heterogeneity of modalities, sensor noise, and environmental uncertainties~\cite{hao2025robust,tang2023hdmap,zhang2025nava,tang2025roboafford,xiao2025team,zhang2025team,hao2025roboafford++,team2025robobrain,liu2025toponav,hao2025mimo,zhang2025your,zhang2025socialnav}.

Current methods for multi-modal HD map prediction often combine image and LiDAR features in BEV space using feature concatenation or attention-based aggregation. These approaches overlook two key factors: (i) the complementary but modality-specific nature of image and LiDAR data, and (ii) the dynamic reliability of these modalities over space and time~\cite{dong2024superfusion}. For example, while camera features excel at capturing lane markings, they fail under poor lighting, and LiDAR offers stable geometry but is limited by sparsity and occlusion. Without modeling these discrepancies, conventional fusion methods can lead to unreliable predictions when one modality is degraded~\cite{hao2025robust}.

To tackle the challenges of multimodal fusion and robustness under degraded conditions, we propose \textbf{\textit{SEF-MAP}}, a Subspace-Expert Fusion framework. The central idea is to capture the unique strengths of each modality while reducing redundancy and enhancing complementarity. To this end, we explicitly decompose BEV features into four subspaces: LiDAR-private, Image-private, Shared, and Interaction. Each subspace is assigned to a specialized expert, which allows the model to retain modality-specific information while also learning cross-modal consensus. To integrate the outputs of these experts, we design an uncertainty-aware gating mechanism that dynamically adjusts expert contributions based on their estimated reliability, enabling more confident experts to play a greater role in the final prediction. To further improve robustness under degraded or missing modalities, we introduce distribution-aware masking during training: when a modality is dropped, its features are replaced by realistic surrogate samples derived from EMA statistics. This encourages each expert to specialize in its designated role and equips the framework to handle partial or corrupted observations. Finally, to stabilize the learning process and avoid expert collapse, we incorporate specialization losses that explicitly enforce the separation of modality-private, shared, and interaction knowledge. Together, these components make \textbf{\textit{SEF-MAP}} a principled and effective solution for robust multimodal HD map prediction.

Our main contributions are summarized as follows:
\begin{itemize}
    \item We develop a subspace-decomposed fusion framework that explicitly separates multimodal BEV features into LiDAR-private, Image-private, Shared, and Interaction streams, thereby mitigating semantic misalignment across modalities.
    \item We propose a distribution-aware masking strategy coupled with specialization losses to enforce expert roles and improve robustness under unimodal degradation, directly addressing the fragility of fused representations when one modality is unreliable.
    \item We design an uncertainty-aware gating mechanism with balance regularizers, enabling adaptive expert selection while preventing redundancy and expert collapse, which alleviates the issue of redundancy and suppression between modalities.
    \item We validate our framework on nuScenes and Argoverse2 and report consistent improvements over state-of-the-art baselines, with +4.2\% mAP on nuScenes and +4.8\% on Argoverse2.
\end{itemize}

\section{RELATED WORK}

\subsection{HD Map Construction Techniques}
HD maps provide crucial prior environmental information for autonomous driving systems. 
The core task of HD map construction is to predict a collection of vectorized and topology-preserving static map elements 
(e.g., lane dividers, road boundaries, pedestrian crossings) in BEV space~\cite{TopoMaskV2,zhang2024hybrimap,zhou2024himap,qiao2023machmap}. 
HD map construction approaches can be categorized into three main types: 
\textbf{Camera-based methods}, \textbf{LiDAR-based methods}, and \textbf{Fusion-based methods}. 
Camera-based methods~\cite{qiao2023bezier,peng2023bevsegformer,qin2022uniformer,zhang2022beverse} 
leverage geometric priors to project image features into BEV space, often introducing spatial distortions. 
These methods typically rely on high-resolution imagery and large-scale pretrained models to improve accuracy~\cite{li2022bevformer,yang2022bevformerv2,zhou2025supermapnet}, 
limiting their practical applicability~\cite{hao2024mapdistill,song2025memfusionmap}. 
LiDAR-based methods~\cite{liu2023vectormapnet,zhou2023lidarmapping,ali2024hdmap} 
provide precise geometric measurements for unified BEV representations but are constrained by point sparsity and environmental interference, 
compromising perception reliability~\cite{dong2024superfusion,hao2025safemap}.
Fusion-based methods attempt to combine the complementary strengths of both modalities through feature concatenation or attention-based aggregation~\cite{MBFusion}. However, existing fusion approaches often suffer from modality misalignment and fail to adequately handle scenarios where one modality is degraded or unreliable, leading to suboptimal performance under challenging conditions~\cite{hao2025safemap}.
Our approach tackles these challenges by decomposing multimodal features into specialized subspaces that preserve modality-specific information while enabling robust cross-modal fusion under degraded conditions.

\subsection{Multimodal Fusion Techniques}
Multimodal fusion, which integrates heterogeneous sensor data such as cameras and LiDAR, has become central to autonomous driving perception~\cite{chen2020hgmf,zhang2024surveyfusion,kong2026robosense,zhang2025humanoidpano,zhang2025video,zheng2025railway,cheng2025exploring,li2025vquala,zhang2025lips,zheng2025generation,wang2025cim,jiang2024recursive,zhang2026you,zhang2024multi,zhang2024trihelper,zhang2025mapnav,wu2025evaluating,gong2025stairway}. Early approaches such as MV3D and AVOD~\cite{chen2017mv3d,zhao2019continuous} relied on direct feature concatenation. While simple, these methods failed to capture meaningful cross-modal correlations and degraded severely under sensor failures~\cite{xiao2024mim,han2025survey}. To overcome these shortcomings, more advanced designs emerged. Transformer-based methods like TransFusion~\cite{bai2022transfusion} employ soft associations between modalities, enabling flexible alignment but with high computational cost. BEV-centric frameworks such as BEVFusion~\cite{liu2023bevfusion,jiang2023semanticbevfusion,liang2022bevfusion} project multi-modal features into a unified BEV space and adaptively combine them through gating networks, improving robustness but incurring significant data and compute demands. Lightweight approaches, exemplified by SparseFusion~\cite{xie2023sparsefusion}, attempt to balance accuracy and efficiency by leveraging sparse attention mechanisms.  

In the context of HD map construction, recent methods have further emphasized temporal reasoning and cross-modal interactions. MemFusionMap~\cite{song2025memfusionmap} introduces memory modules to strengthen sequential modeling, while InteractionMap~\cite{wu2025interactionmap} enhances fine-grained feature exchanges across modalities. Despite their improvements, these approaches remain computationally intensive and still lack explicit modeling of heterogeneous modality-specific cues. Overall, the development of multimodal fusion has evolved from naive concatenation to transformer-driven, BEV-aligned, and memory-augmented strategies, yet cross-modal misalignment remains insufficiently addressed. This gap motivates our subspace-expert fusion framework, which explicitly disentangles modality-private, shared, and interaction-specific subspaces to achieve more robust and interpretable multi-modal HD map prediction.

\section{METHODOLOGY}
\subsection{Problem Formulation and Method Overview}
\label{subseq:problemnotation}
\textbf{Problem Formulation.} Given synchronized multi-view images $I=\{I_c\}_{c=1}^{N_{\mathrm{cam}}}$ and a LiDAR point cloud $L$, our goal is to predict an HD map $\hat{\mathbf{Y}}$ in a BEV frame. Let modality encoders produce BEV features $F_I=E_I(I)\in\mathbb{R}^{H\times W\times C}$ and $F_L=E_L(L)\in\mathbb{R}^{H\times W\times C}$. Our method introduces a fusion module $\mathcal{F}_\theta$ (detailed in Sec.~\ref{subseq:subspace}) that operates purely in BEV and is followed by task heads $Dec$ to decode vectorized map elements. The learning objective on dataset $\mathcal{D}=\{(I^{(n)},L^{(n)},\mathbf{Y}^{(n)})\}_{n=1}^{N}$ is
\begin{equation}
\label{eq:objective}
\min_{\theta,\phi}\;\frac{1}{N}\sum_{n=1}^{N}
\mathcal{L}_{\mathrm{task}}\!\left(
Dec_{\phi}\!\big(\mathcal{F}_\theta(F_I^{(n)},F_L^{(n)})\big),
\;\mathbf{Y}^{(n)}
\right),
\end{equation}
where $\mathcal{L}_{\mathrm{task}}$ is the standard set-based classification and geometric regression loss used for HD map elements.

\begin{figure*}[t!]
    \centering
    \includegraphics[width=0.88\textwidth]{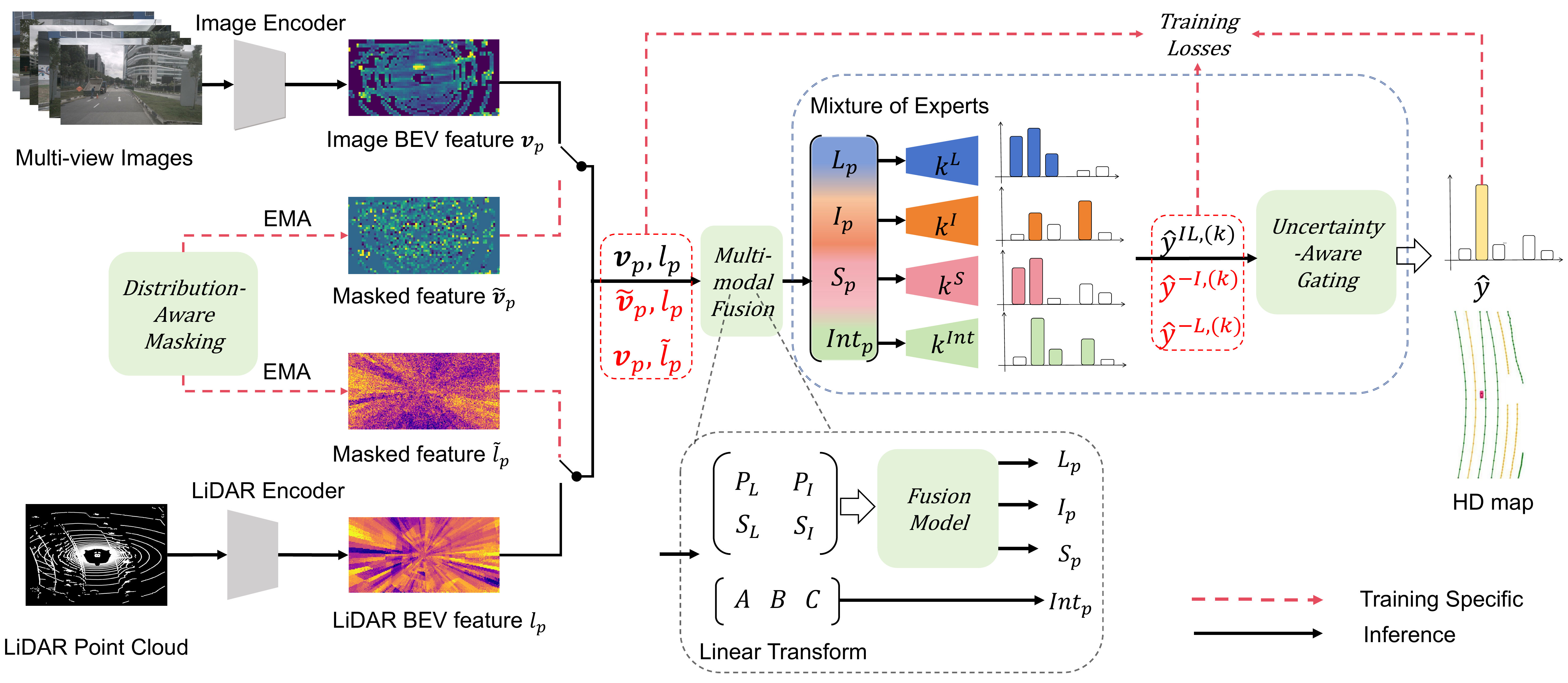} % 图片文件
    \caption{\textbf{Overall architecture of SEF-MAP framework.} The system takes multi-view images and LiDAR point clouds as inputs, which are encoded into BEV features through respective encoders. These features are then decomposed into four semantic subspaces (LiDAR-private, Image-private, Shared, and Interaction) via linear transformations. During training, distribution-aware masking creates surrogate features using EMA statistics to simulate modality degradation scenarios, with specialization losses enforcing expert roles. During inference, only the intact forward pass is performed without masking. A mixture of experts processes each subspace, and an uncertainty-aware gating mechanism dynamically weights expert outputs based on their predicted variance to generate the final HD map prediction.}
    \label{fig:overall}
\vspace{-1.5em}
\end{figure*}

\textbf{Method Overview.} The overall architecture of our method is shwon in Fig.~\ref{fig:overall}. To enable precise and adaptive cross-modal fusion, we decompose BEV features into four subspaces: LiDAR-private, Image-private, Shared, and Interaction spaces. A cell-wise, uncertainty-aware gating module then reweights these experts to form the final prediction. During training, we clarify expert roles and improve single-modality behavior by replacing one modality with a surrogate sampled from its empirical BEV feature distribution, creating realistic masked variants without disrupting feature statistics. Learning is further stabilized by a lightweight balance regularizer that prevents expert starvation. At inference, a single intact forward aggregates expert outputs via uncertainty-aware gating, without masking or auxiliary regularizers.

\subsection{Subspace-Decomposed Multimodal Fusion}
\label{subseq:subspace}

For each BEV cell $p \in \mathcal{P}$, we fuse the LiDAR BEV feature $\boldsymbol{\ell}_p \in \mathbb{R}^{C}$ and the image BEV feature $\mathbf{v}_p \in \mathbb{R}^{C}$ by projecting them into four semantically distinct subspaces: \emph{LiDAR} ($\mathbf{L}_p$), \emph{Image} ($\mathbf{I}_p$), \emph{Shared} ($\mathbf{S}_p$), and \emph{Interaction} ($\mathbf{Int}_p$).  
With learnable linear maps $\mathbf{P}_L,\mathbf{P}_I,\mathbf{S}_L,\mathbf{S}_I \in \mathbb{R}^{C\times C}$ and a low-rank bilinear form parameterized by $\mathbf{A},\mathbf{B} \in \mathbb{R}^{R\times C}$ and $\mathbf{C} \in \mathbb{R}^{C\times R}$ with $R \ll C$,where $C$ is the BEV channel width, $R$ is the interaction rank.  we first compute modality-private and shared projections:
\begin{equation}
\label{eq:private-compact}
\begin{aligned}
\mathbf{u}^{L}_p &= \mathbf{P}_L\,\boldsymbol{\ell}_p, \quad
\mathbf{u}^{I}_p = \mathbf{P}_I\,\mathbf{v}_p, \\
\mathbf{r}^{L}_p &= \mathbf{S}_L\,\boldsymbol{\ell}_p, \quad
\mathbf{r}^{I}_p = \mathbf{S}_I\,\mathbf{v}_p,
\end{aligned}
\end{equation}
where $\mathbf{u}^{L}_p$ preserves LiDAR-specific geometric cues such as range and 3D structure that remain stable under illumination changes or low-light conditions, while $\mathbf{u}^{I}_p$ retains image-specific semantic cues such as appearance and road markings. The features $\mathbf{r}^{L}_p$ and $\mathbf{r}^{I}_p$ represent modality-shared evidence that is consistently observable across both LiDAR and camera, such as lane continuity.

To regulate subspace features, we introduce an auxiliary \emph{space loss} defined as
\begin{equation}
\label{eq:space}
\mathcal{L}_{\mathrm{space}}=\frac{1}{|\mathcal{P}|}\sum_{p\in\mathcal{P}}\Big(\mathcal{L}_{\mathrm{uni}}+\mathcal{L}_{\mathrm{shr}}+\mathcal{L}_{\mathrm{int}}\Big),
\end{equation}
where three components operate at each BEV cell $p$.  \\
The \emph{private decorrelation loss} $\mathcal{L}_{\mathrm{uni}}$ enforces independence between private and opposite-modality features using HSIC:
\begin{equation}
\mathcal{L}_{\mathrm{uni}}=\mathrm{HSIC}(u^L_p, v_p)+\mathrm{HSIC}(u^I_p, \ell_p).
\end{equation}
The \emph{shared alignment loss} $\mathcal{L}_{\mathrm{shr}}$ encourages consistency between the shared projections $r^L_p$ and $r^I_p$ via negative correlation:
\begin{equation}
\mathcal{L}_{\mathrm{shr}}=-\mathrm{corr}(r^L_p, r^I_p).
\end{equation}
The \emph{interactive contrastive loss} $\mathcal{L}_{\mathrm{int}}$ strengthens the cross-modal sensitivity of the interaction subspace $\mathbf{Int}_p$ through InfoNCE:
\begin{equation}
\label{eq:int}
\mathcal{L}_{\mathrm{int}}=-\log\frac{\exp\phi(\mathbf{Int}_p,\ell_p,v_p)}{\sum_{p'}\exp\phi(\mathbf{Int}_p,\ell_{p'},v_{p'})}.
\end{equation}

Together, these regularizers disentangle private, shared, and interaction subspaces, thereby improving modality robustness and cross-modal complementarity.

Based on these projections, four subspaces can be constructed as:
\begin{equation}
\label{eq:subspaces-compact}
\begin{aligned}
\mathbf{L}_p   &= f^{L}_{\text{fusion}}\!\big([\mathbf{u}^{L}_p \,\|\, \mathbf{r}^{I}_p]\big), \\
\mathbf{I}_p   &= f^{I}_{\text{fusion}}\!\big([\mathbf{r}^{L}_p \,\|\, \mathbf{u}^{I}_p]\big), \\
\mathbf{S}_p   &= f^{S}_{\text{fusion}}\!\big([\mathbf{r}^{L}_p \,\|\, \mathbf{r}^{I}_p]\big), \\
\mathbf{Int}_p &= \mathbf{C}\!\left[\big(\mathbf{A}\boldsymbol{\ell}_p\big)\odot\big(\mathbf{B}\mathbf{v}_p\big)\right].
\end{aligned}
\end{equation}
where $\odot$ denotes element-wise multiplication. In this formulation, $\mathbf{L}_p$ and $\mathbf{I}_p$ serve as the modality-inclined information, $\mathbf{S}_p$ encodes common and modality-invariant information, and $\mathbf{Int}_p$ captures complementary cross-modal interactions through multiplicative gating. The interaction space is particularly useful for resolving ambiguities caused by occlusions or weak signals in a single modality. For convenience, we denote $\mathbf{z}_p^{(k)} \in \{\mathbf{L}_p, \mathbf{I}_p, \mathbf{S}_p, \mathbf{Int}_p\}$ as the subspace features assigned to each expert $k \in K$, where $K=\{\mathbf{k}^{L}, \mathbf{k}^{I}, \mathbf{k}^{S}, \mathbf{k}^{Int}\}$. All linear maps are implemented as $1{\times}1$ BEV convolutions and operate independently at each BEV cell $p$.

\subsection{Distribution-Aware Masking with Specialization Losses}

Since either modality may be degraded by noise, occlusion, or sensor perturbations, we propose a \emph{distribution-aware masked forward} strategy to encourage clear expert specialization and robustness to partial observations. To this end, we maintain per-channel means and variances of the BEV features, denoted as $(\boldsymbol{\mu}_I,\boldsymbol{\sigma}_I^2)$ for images and $(\boldsymbol{\mu}_L,\boldsymbol{\sigma}_L^2)$ for LiDAR, which are updated as exponential moving averages (EMA) during training to provide stable distribution estimates. When masking occurs, the suppressed modality is replaced with a surrogate feature sampled from its empirical distribution:
\begin{equation}
\label{eq:distaware}
\tilde{\mathbf{v}}_p \sim \mathcal{N}\!\big(\boldsymbol{\mu}_I,\operatorname{diag}(\boldsymbol{\sigma}_I^2)\big), \qquad
\tilde{\boldsymbol{\ell}}_p \sim \mathcal{N}\!\big(\boldsymbol{\mu}_L,\operatorname{diag}(\boldsymbol{\sigma}_L^2)\big).
\end{equation}

During training, three types of forward passes are performed.  
The intact pass $\hat{\mathbf{y}}^{IL}$ uses both modalities $(\boldsymbol{\ell}, \mathbf{v})$, the image-masked pass $\hat{\mathbf{y}}^{-I}$ uses LiDAR together with surrogate image features $(\boldsymbol{\ell}, \tilde{\mathbf{v}})$, and the LiDAR-masked pass $\hat{\mathbf{y}}^{-L}$ uses camera features with surrogate LiDAR inputs $(\tilde{\boldsymbol{\ell}}, \mathbf{v})$.  
These variants approximate unimodal conditions: private experts are encouraged to align with their respective modality, the shared expert is trained to remain consistent across both, and the interaction expert learns cues that weaken when either modality is missing.  
By sampling from empirical marginal distributions, the masked inputs preserve channel energy and normalization statistics, thereby avoiding out-of-distribution artifacts and providing informative gradients for optimization. At inference, only the intact forward pass is employed without masking.

To translate the behavior of masked passes into explicit learning signals, we design \emph{specialization losses}. 
Specifically, we define per-cell losses that contrast intact and masked predictions, thereby clarifying the role of each expert. 

We denote by $\hat{\mathbf{y}}^{IL,(k)}$, $\hat{\mathbf{y}}^{-I,(k)}$, and $\hat{\mathbf{y}}^{-L,(k)}$ the outputs of expert $k$ under intact, image-masked, and LiDAR-masked conditions, respectively. 
Here, $d(\cdot,\cdot)$ is a task-consistent dissimilarity, $\gamma\in(0,1]$ a balance coefficient, and $m>0$ a margin.

% To translate the behavior of these masked passes into explicit learning signals, we design specialization losses.  
% Let $\hat{\mathbf{y}}^{IL,(k)}$, $\hat{\mathbf{y}}^{-I,(k)}$, and $\hat{\mathbf{y}}^{-L,(k)}$ denote predictions from intact, image-masked, and LiDAR-masked forwards of expert $k$.  

% We operationalize this intuition by defining per-cell specialization losses that contrast intact and masked predictions. 
% Let $d(\cdot,\cdot)$ be a task-consistent dissimilarity, $\gamma\in(0,1]$ a balance coefficient, and $m>0$ a margin.  
For the LiDAR-private expert:
\begin{equation}
\label{eq:spec-lidar}
\mathcal{L}^{L}_p=d\!\big(\hat{\mathbf{y}}^{IL,(L)}_p,\hat{\mathbf{y}}^{-I,(L)}_p\big)-\gamma\,d\!\big(\hat{\mathbf{y}}^{IL,(L)}_p,\hat{\mathbf{y}}^{-L,(L)}_p\big),
\end{equation}
while for the Image-private expert:
\begin{equation}
\label{eq:spec-image}
\mathcal{L}^{I}_p=d\!\big(\hat{\mathbf{y}}^{IL,(I)}_p,\hat{\mathbf{y}}^{-L,(I)}_p\big)-\gamma\,d\!\big(\hat{\mathbf{y}}^{IL,(I)}_p,\hat{\mathbf{y}}^{-I,(I)}_p\big).
\end{equation}
The Shared expert is optimized to remain consistent under both maskings:
\begin{equation}
\label{eq:spec-shared}
\mathcal{L}^{S}_p=d\!\big(\hat{\mathbf{y}}^{IL,(S)}_p,\hat{\mathbf{y}}^{-I,(S)}_p\big)+d\!\big(\hat{\mathbf{y}}^{IL,(S)}_p,\hat{\mathbf{y}}^{-L,(S)}_p\big),
\end{equation}
and the Interaction expert emphasizes cross-modal complementarity:
\begin{equation}
\begin{aligned}
\mathcal{L}^{Int}_p &= \big[m-d\!\big(\hat{\mathbf{y}}^{IL,(Int)}_p,\hat{\mathbf{y}}^{-I,(Int)}_p\big)\big]_+ \\
&\quad + \big[m-d\!\big(\hat{\mathbf{y}}^{IL,(Int)}_p,\hat{\mathbf{y}}^{-L,(Int)}_p\big)\big]_+.
\end{aligned}
\end{equation}

Averaging across cells yields
\begin{equation}
\label{eq:exp}
\mathcal{L}_{\mathrm{spec}}=\frac{1}{|\mathcal{P}|}\sum_{p\in\mathcal{P}}\Big(\mathcal{L}^{L}_p+\mathcal{L}^{I}_p+\mathcal{L}^{S}_p+\mathcal{L}^{Int}_p\Big).
\end{equation}

\begin{table*}[!ht]
\centering
\caption{\textbf{Comparisons with state-of-the-art methods on nuScenes val set.} 
Our method outperforms all existing approaches in both single-class APs and overall mAP.}
\label{tab:sota_nuscenes}
\resizebox{0.88\textwidth}{!}{
\begin{tabular}{lcccccccc}
\toprule
\textbf{Method} & \textbf{Modality} & \textbf{Backbone} & \textbf{Epochs} & $AP_{ped}$ & $AP_{divider}$ & $AP_{boundary}$ & \textbf{mAP} \\
\midrule
HDMapNet~\cite{hdmapnet}   & C          & E-B0        & 30  & 14.4 & 21.7 & 33.0 & 23.0 \\
HDMapNet~\cite{hdmapnet}   & L          & PointPillars & 30  & 10.4 & 24.1 & 37.9 & 24.1 \\
HDMapNet~\cite{hdmapnet}   & C\&L       & E-B0 \& PointPillars & 30  & 16.3 & 29.6 & 46.7 & 31.0 \\
\midrule
VectorMapNet~\cite{vectormapnet} & C    & R50         & 110 & 36.1 & 47.3 & 39.3 & 40.9 \\
VectorMapNet~\cite{vectormapnet} & L    & PointPillars & 110 & 25.7 & 37.6 & 38.6 & 34.0 \\
VectorMapNet~\cite{vectormapnet} & C\&L & R50 \& PointPillars & 110 & 37.6 & 50.5 & 47.5 & 45.2 \\
\midrule
MapTR~\cite{maptr} & C       & R50        & 24  & 46.3 & 51.5 & 53.1 & 50.3 \\
MapTR~\cite{maptr} & L       & SECOND     & 24  & 48.5 & 53.7 & 64.7 & 55.6 \\
MapTR~\cite{maptr} & C\&L    & R50 \& SECOND & 24  & 55.9 & 62.3 & 69.3 & 62.5 \\
\midrule
\rowcolor{cyan!10} 
\textbf{SEF-Map (Ours)} & C\&L & R50 \& SECOND & \textbf{24}  
& \textbf{61.6} & \textbf{66.7} & \textbf{71.8} & \textbf{66.7(+4.2)} \\
\bottomrule
\end{tabular}}
\end{table*}

\begin{table*}[t]
\centering
\caption{\textbf{Results on Argoverse2 dataset.} $\dagger$ denotes our re-implementation following the setting in the paper.}
\label{tab:argo_results}
\resizebox{0.88\textwidth}{!}{
\setlength{\tabcolsep}{4pt} % 调整列间距
\renewcommand{\arraystretch}{1.1} % 调整行距
\begin{tabular}{lcccccccc}
\toprule
\textbf{Method} & \textbf{Modality} & \textbf{Backbone} & \textbf{Epochs} & $AP_{ped}$ & $AP_{divider}$ & $AP_{boundary}$ & \textbf{mAP} \\
\midrule
HDMapNet~\cite{hdmapnet}   & C     & E-B0        & 30  & 13.1 &  5.7 & 37.6 & 18.8 \\
VectorMapNet~\cite{vectormapnet} & C & R50         & 110 & 38.3 & 36.1 & 39.2 & 37.9 \\
MapTR$^{\dagger}$~\cite{maptr}   & C     & R50        & 6   & 58.7 & 59.3 & 60.3 & 59.4 \\
MapTR$^{\dagger}$~\cite{maptr}   & C\&L  & R50 \& SECOND & 6   & 65.1 & 61.6 & 75.1 & 67.3 \\
\midrule
\rowcolor{cyan!10} 
\textbf{SEF-Map (Ours)} & C\&L & R50 \& SECOND & \textbf{6}   
& \textbf{70.7} & \textbf{66.7} & \textbf{78.8} & \textbf{72.1(+4.8)} \\
\bottomrule
\end{tabular}}
\end{table*}

% \vspace{-10em}
\subsection{Uncertainty-Aware Gating and Stabilization}
\label{subseq:uncertainty}
Subspace reliability varies across BEV cells, so the model should prefer the expert that is both informative and confident at each location. For cell $p\in\mathcal{P}$ and expert $k\in K$, the head $H_k$ maps the subspace feature $\mathbf{z}_p^{(k)}$ to a mean prediction $\boldsymbol{\mu}_p^{(k)}\!\in\!\mathbb{R}^{D}$ and a variance $\boldsymbol{\sigma}^{2\,(k)}_{p}\!\in\!\mathbb{R}^{D}$. Here the variance is parameterized as $\log\sigma^2$ for stability. A lightweight gate $g$ produces logits $\alpha_p^{(k)}\!\in\!\mathbb{R}$. We down-weight uncertain experts by subtracting their average predicted variance from the logits:
\begin{equation}
\label{eq:gate}
w_p^{(k)}=\operatorname{softmax}_{k}\!\Big(\alpha_p^{(k)}-\beta\,\bar{\sigma}^{2\,(k)}_{p}\Big),\quad
\bar{\sigma}^{2\,(k)}_{p}=\tfrac{1}{D}\sum_{d=1}^{D}\sigma^{2\,(k)}_{p,d},
\end{equation}
where $\beta>0$ controls the penalty strength and $D$ is the head output dimension. The final per-cell prediction is
\begin{equation}
\label{eq:mixture}
\hat{\mathbf{y}}_{p}=\sum_{k\in K} w_p^{(k)}\,\boldsymbol{\mu}_p^{(k)}\in\mathbb{R}^{D}.
\end{equation}
% Subspaces should provide complementary evidence rather than duplicating one another. To reduce leakage and improve identifiability, we introduce the \emph{Orthogonality Loss}, which penalizes pairwise correlations among normalized subspace features:
% \begin{equation}
% \label{eq:orth}
% \Omega_{\mathrm{orth}}=\frac{1}{|\mathcal{P}|}\sum_{p\in\mathcal{P}}\sum_{\substack{k,k'\in K\\k\neq k'}} 
% \left(
% \frac{{\mathbf{z}_p^{(k)}}^{\!\top}\mathbf{z}_p^{(k')}}{\lVert \mathbf{z}_p^{(k)}\rVert_2\,\lVert \mathbf{z}_p^{(k')}\rVert_2}
% \right)^{\!2}.
% \end{equation}
Moreover, gating can collapse onto a single expert if left unconstrained, which is called expert starvation and harms specialization. Therefore, we regularize the average usage of each expert toward a uniform prior $1/|K|$,
\begin{equation}
\label{eq:balance}
\Omega_{\mathrm{bal}}=\sum_{k\in K}\left(\bar{w}_k-\tfrac{1}{|K|}\right)^{2},
\qquad
\bar{w}_k=\frac{1}{|\mathcal{P}|}\sum_{p\in\mathcal{P}} w_p^{(k)}.
\end{equation}
Together, Eqs.~\eqref{eq:gate}–\eqref{eq:balance} yield gates that are cell-wise adaptive, variance-aware, and resistant to expert collapse.

\begin{algorithm}[t]
\caption{Training and Inference for Subspace-Decomposed Fusion}
\label{alg:train-infer}
\small
\Input{Batch $\{(I,L,\mathbf{Y})\}$; encoders $E_I,E_L$; lifts $T_I,T_L$; maps $\mathbf{P}_\star,\mathbf{S}_\star,\mathbf{A},\mathbf{B},\mathbf{C}$; heads $\{H_k\}_{k\in K}$; gate $g$; EMA stats $(\boldsymbol{\mu}_I,\boldsymbol{\sigma}_I^2),(\boldsymbol{\mu}_L,\boldsymbol{\sigma}_L^2)$}
\Output{$\hat{\mathbf{y}}$}

\Proc{TrainBatch}{
$F_I \gets T_I(E_I(I));\quad F_L \gets T_L(E_L(L))$\;
\For{$p\in\mathcal{P}$}{
  $\mathbf{u}^L_p,\mathbf{u}^I_p,\mathbf{u}^{S}_p,\mathbf{u}^{Int}_p \gets$ Eq.~\eqref{eq:subspaces-compact}; set $\mathbf{z}^{(k)}_p$\;
  $(\boldsymbol{\mu}^{(k)}_p,\boldsymbol{\sigma}^{2\,(k)}_p) \gets H_k(\mathbf{z}^{(k)}_p)$ for $k\in K$\;
}
$\{\alpha^{(k)}_p\}\gets g(\{\mathbf{z}^{(k)}_p\}_k)$\;
$w^{(k)}_p \gets$ Eq.~\eqref{eq:gate}\;
$\hat{\mathbf{y}}^{IL}\gets$ Eq.~\eqref{eq:mixture}\;
Sample $\tilde{\mathbf{v}}_p,\tilde{\boldsymbol{\ell}}_p$ via Eq.~\eqref{eq:distaware} \Comment{compute $\hat{\mathbf{y}}^{-I},\hat{\mathbf{y}}^{-L}$}\;
Compute $\mathcal{L}$ using Eqs. \eqref{eq:space}--\eqref{eq:int}, \eqref{eq:spec-lidar}--\eqref{eq:exp},and \eqref{eq:balance}--\eqref{eq:overall-loss} \Comment{update model parameters and EMA stats}\;
}

\Proc{Inference{$I,L$}}{
$F_I \gets T_I(E_I(I));\quad F_L \gets T_L(E_L(L))$\;
\For{$p\in\mathcal{P}$}{
  $\mathbf{z}^{(k)}_p \gets$ Eq.~\eqref{eq:subspaces-compact}; \ $\boldsymbol{\mu}^{(k)}_p \gets H_k(\mathbf{z}^{(k)}_p)$ for $k\in K$\;
}
$w^{(k)}_p \gets$ Eq.~\eqref{eq:gate}$;$ \;
\Return $\hat{\mathbf{y}}=\sum_{k\in K} w^{(k)}_p\,\boldsymbol{\mu}^{(k)}_p$ \Comment{Eq.~\eqref{eq:mixture}}\;
}
\end{algorithm}

\subsection{Training and Inference Pipeline}
\label{subseq:training}

\textbf{Loss Definition.} Aggregating per-cell predictions yields the set-level output $\hat{\mathbf{Y}}^{IL}$. The task loss compares intact predictions against ground truth:
\begin{equation}
\label{eq:task}
\mathcal{L}_{\mathrm{task}}=\mathcal{L}_{\mathrm{task}}\!\big(\hat{\mathbf{Y}}^{IL},\mathbf{Y}\big).
\end{equation}

Finally, the complete training objective integrates all components:
\begin{equation}
\label{eq:overall-loss}
\begin{aligned}
\mathcal{L}=&\ \mathcal{L}_{\mathrm{task}}
+\lambda_{\mathrm{spec}}\mathcal{L}_{\mathrm{spec}}
+\lambda_{\mathrm{space}}\mathcal{L}_{\mathrm{space}}
+\lambda_{\mathrm{bal}}\Omega_{\mathrm{bal}},
\end{aligned}
\end{equation}
where $\lambda_{\mathrm{spec}}, \lambda_{\mathrm{space}}, \lambda_{\mathrm{bal}}>0$ balance the contributions of different terms.

\textbf{Training.} Each mini-batch mixes intact and masked conditions so the gate learns when to trust private, shared, or interaction evidence. Inputs are encoded and lifted to BEV. For each cell we compute the four subspace features using Eq.~\eqref{eq:subspaces-compact}, and the expert heads return mean and variance estimates. The gate produces logits and we apply the variance penalty in Eq.~\eqref{eq:gate} to obtain weights; the intact prediction then follows from Eq.~\eqref{eq:mixture}. Surrogate features are sampled with Eq.~\eqref{eq:distaware} to create two masked passes, and the same forward path yields their predictions. We evaluate the objective in Eq.~\eqref{eq:overall-loss} and finally update network parameters by backpropagation and refresh the exponential moving averages used for distribution-aware masking.

\textbf{Inference.} At test time, we perform a single intact forward pass without any masking or auxiliary objectives. For each BEV cell, the model first computes the corresponding subspace features, which are then processed by the expert heads to produce both mean predictions and associated variances. The gating module subsequently generates weights according to Eq.~\eqref{eq:gate}, allowing reliable experts to contribute more strongly. The final output is obtained as a weighted mixture of expert predictions, as defined in Eq.~\eqref{eq:mixture}. Since no additional masking or auxiliary losses are involved during inference, the runtime remains comparable to that of a standard BEV fusion network.

\newcommand{\cmark}{\checkmark}
\newcommand{\xmark}{$\times$}

\section{EXPERIMENTS}

\subsection{Experimental Setup}

\textbf{Dataset.} We conducted experiments on the nuScenes and Argoverse2 datasets. 
The nuScenes dataset contains 1000 driving sequences, each with six camera streams and LiDAR sweeps annotated at 2Hz. 
We follow the preprocessing pipeline in MapTR~\cite{maptr} to ensure consistent data preparation. 
In our experiments, we focus on three representative map elements: pedestrian crossings, lane dividers, and road boundaries. 
The Argoverse2 dataset also provides 1000 sequences, partitioned into 700 for training, 150 for validation, and 150 for testing. 
Each sequence spans 15 seconds, with synchronized 20Hz camera images and 10Hz LiDAR sweeps. 
For both datasets, we apply spatial alignment, temporal synchronization, and feature normalization to enable reliable multi-modal fusion during training. 
Owing to their diverse urban and suburban scenes, these datasets serve as strong benchmarks for evaluating the generalization ability of our method.

\textbf{Evaluation Metrics.} We evaluate the quality of the predicted HD maps using average precision (AP). To assess the alignment between predicted and ground-truth map elements, we compute the Chamfer distance (DChamfer) at multiple thresholds (\(\tau \in T\), where \(T = \{0.5, 1.0, 1.5\}\) meters). The final AP score is averaged over all these thresholds, providing a comprehensive measure of the model's performance across different levels of spatial accuracy.

% ===== 表格正文 =====

\textbf{Implementation Details.} We train \textbf\textit{{SEF-MAP}} on 8 NVIDIA RTX 4090 GPUs using the AdamW optimizer. The learning rate is set to 6e-4 with step decay, and the mini-batch size is 16. The backbone uses ResNet50 for image features and SECOND for LiDAR features. The overall loss function consists of task loss, space loss, specialization loss, and balance regularization, with weights \(\lambda_{\text{task}} = 1.0\), \(\lambda_{\text{space}} = 0.1\), \(\lambda_{\text{spec}} = 0.5\), and \(\lambda_{\text{bal}} = 0.05\).  To ensure consistency, the sampled forward passes adopt exactly the same input configuration as the intact inputs, so that the gating mechanism learns under comparable conditions.

\begin{figure*}[!t]
    \centering
    \includegraphics[width=0.91\textwidth]{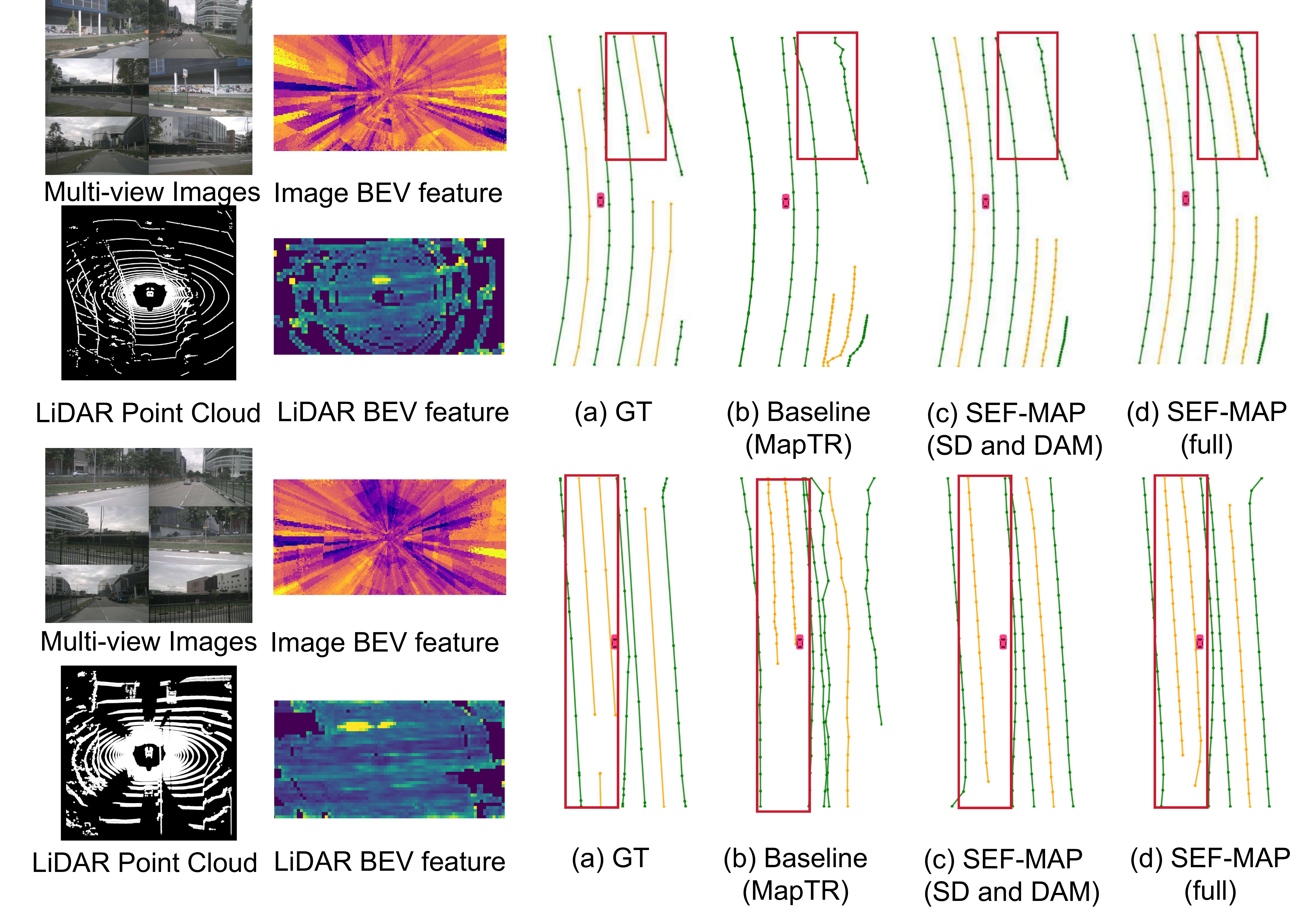} % 图片文件
    \vspace{-0.5em}
    \caption{\textbf{Qualitative results on nuScenes.} We present two sample scenes from nuScenes: (a) Ground Truth. (b) Baseline (MapTR). (c) SEF-MAP (SD and DAM). (d) \textbf{SEF-MAP (full).} The red boxes highlight challenging regions where our method demonstrates significant improvements over the baseline. In these highlighted areas, the baseline MapTR produces noticeable errors and incomplete map predictions. The full SEF-MAP model with uncertainty-aware gating (d) achieves the most precise vectorized map reconstruction, particularly in the challenging regions marked by red boxes, demonstrating the effectiveness of our subspace decomposition and adaptive fusion strategy.  }
    \label{fig:quali}

\vspace{-1.6em}
\end{figure*}

\subsection{Comparison with SOTA Methods}

% We compare \textbf{\textit{SEF-MAP}} with several state-of-the-art models, including HDMapNet \cite{hdmapnet}, VectorMapNet \cite{vectormapnet}, and MapTR \cite{maptr}, on the nuScenes and Argoverse2 datasets, as shown in Tab.~\ref{tab:sota_nuscenes} and Tab.~\ref{tab:argo_results}. \textbf{\textit{SEF-MAP}} consistently outperforms all baselines: on nuScenes, the most notable improvement is a +5.7\% gain in \(AP_{\text{ped}}\) compared to MapTR, while on Argoverse2, \textbf{\textit{SEF-MAP}} achieves the largest boost in overall mAP with +4.8\%. These results highlight the effectiveness of our subspace decomposition and uncertainty-aware gating modules and further indicate that the superiority of multi-modal designs over single-modality counterparts is systematic rather than incidental, underscoring the importance of fully exploiting modality-specific representations to guide the learning process for robust HD map construction.
We compare \textbf{\textit{SEF-MAP}} with several state-of-the-art models, including HDMapNet \cite{hdmapnet}, VectorMapNet \cite{vectormapnet}, and MapTR \cite{maptr}, on the nuScenes and Argoverse2 datasets, as shown in Tab.~\ref{tab:sota_nuscenes} and Tab.~\ref{tab:argo_results}. \textbf{\textit{SEF-MAP}} consistently outperforms all baselines across both datasets and all map element categories: on nuScenes, the most notable improvement is a +5.7\% gain in \(AP_{\text{ped}}\) compared to MapTR, along with substantial improvements of +4.4\% in \(AP_{\text{divider}}\) and +2.5\% in \(AP_{\text{boundary}}\), resulting in an overall mAP improvement of +4.2\%. On Argoverse2, \textbf{\textit{SEF-MAP}} achieves even more significant gains, with the largest boost in overall mAP of +4.8\%, demonstrating strong generalization capability across different datasets and driving scenarios. Notably, the consistent improvements across all three map element types (pedestrian crossings, lane dividers, and road boundaries) indicate that our framework effectively captures both fine-grained semantic details and geometric structures. These results highlight the effectiveness of our subspace decomposition and uncertainty-aware gating modules in handling diverse environmental conditions and sensor configurations. The substantial performance gains further indicate that the superiority of multi-modal designs over single-modality counterparts is systematic rather than incidental, underscoring the importance of fully exploiting modality-specific representations while maintaining robust cross-modal interactions to guide the learning process for reliable HD map construction.

\begin{table}[t]
\centering
\caption{\textbf{Ablation on subspace decomposition (SD), distribution-aware masking (DAM), and uncertainty-aware gating (UAG).}
The full model combining all three components achieves the state-of-the-art performance, outperforming individual modules and pairwise combinations on nuScenes val (24 epochs).}
\label{tab:ablation}
\setlength{\tabcolsep}{3pt}
\renewcommand{\arraystretch}{1.15}
\resizebox{0.48\textwidth}{!}{
\begin{tabular}{ccc c c c c c}
\toprule
\textbf{SD} & \textbf{DAM} & \textbf{UAG} & \textbf{Mod.} &
$AP_{ped}$ & $AP_{divider}$ & $AP_{boundary}$ & \textbf{mAP} \\
\midrule
\xmark & \xmark & \xmark & C\&L & 55.9 & 62.3 & 69.3 & 62.5 \\  % baseline
\cmark & \xmark & \xmark & C\&L & 56.8 & 63.1 & 70.2 & 63.4 \\  % SD only
\xmark & \cmark & \xmark & C\&L & 57.7 & 64.6 & 71.0 & 64.4 \\  % DAM only
\xmark & \xmark & \cmark & C\&L & 56.1 & 62.8 & 69.7 & 62.9 \\  % UAG only
\cmark & \cmark & \xmark & C\&L & 59.1 & 66.7 & 72.0 & 65.9 \\  % SD + DAM
\cmark & \xmark & \cmark & C\&L & 59.2 & 64.9 & 70.7 & 64.9 \\  % SD + UAG
\xmark & \cmark & \cmark & C\&L & 57.8 & 65.3 & 72.2 & 65.1 \\  % DAM + UAG
\rowcolor{cyan!10}
\cmark & \cmark & \cmark & C\&L & \textbf{61.6} & \textbf{66.7} & \textbf{71.8} & \textbf{66.7} \\  % full
\bottomrule
\end{tabular}}
\end{table}

\begin{table}[t]
\centering
\caption{\textbf{Expert group ablation configurations and results}. The full model achieves the best performance, significantly outperforming configurations with only private experts (L+I) or only cross-modal experts (S+Int).}
\label{tab:expert-group}
\setlength{\tabcolsep}{3pt}
\renewcommand{\arraystretch}{1.15}
\resizebox{0.48\textwidth}{!}{
\begin{tabular}{l c c c c}
\toprule
\textbf{Expert Setting} & $AP_{ped}$ & $AP_{divider}$ & $AP_{boundary}$ & \textbf{mAP} \\
\midrule
MapTR~\cite{maptr} (baseline) & 55.9 & 62.3 & 69.3 & 62.5 \\
Only L+I (Private)            & 57.7 & 64.2 & 69.4 & 63.8 \\
Only S+Int (Cross-modal)      & 54.6 & 60.5 & 66.7 & 60.6 \\
\rowcolor{cyan!10}
\textbf{SEF-Map (Full)}                          & \textbf{61.6} & \textbf{66.7} & \textbf{71.8} & \textbf{66.7} \\
\bottomrule
\end{tabular}}
\end{table}

\subsection{Ablation Studies}

\textbf{Analysis on Different Modules.} To systematically evaluate the effectiveness of each module in our proposed model, we train the model by removing each component one at a time, as presented in Tab.~\ref{tab:ablation}. We design the following configurations:
We design the following ablation models: 
(1) \textbf{w/o DAM + w/o UAG}: we remove DAM and replace UAG with an MLP;
(2) \textbf{w/o SD + w/o UAG}: we remove SD and replace UAG with an MLP;
(3) \textbf{w/o SD + w/o DAM}: we remove both SD and DAM modules; 
(4) \textbf{w/o UAG}: we replace \textit{uncertainty-aware gating (UAG)} with an MLP for computing expert weights; 
(5) \textbf{w/o DAM}: we remove the \textit{distribution-aware masking (DAM)} module; 
(6) \textbf{w/o SD}: we remove the \textit{subspace decomposition (SD)} module from the model;

The results show that introducing SD and DAM brings the most substantial gains, with mAP improving by about 3.4\% compared to the baseline. UAG alone yields only a marginal +0.4\% gain, but when combined with SD and DAM it provides an additional +0.8\% improvement, showing that UAG is most effective when enhancing their complementary features. These findings confirm the complementary roles of the three modules, with SD and DAM providing the strongest improvements and UAG refining the fusion process.

% \textbf{Analysis on Expert Group Ablation.} Additionally, we perform an expert group ablation to evaluate the importance of different expert configurations. We design the following configurations:
% \begin{enumerate}

%     (1) \textbf{Private Experts Only (L+I)}: using only the \textit{private experts from LiDAR and Image} results in slightly better performance than the baseline MapTR, but still underperforms compared to the full model, indicating that relying solely on private experts limits the model’s ability to utilize complementary information; 
% (2) \textbf{Cross-Modal Experts Only (S+Int)}: using only the \textit{cross-modal experts} results in even weaker performance, confirming that cross-modal experts alone cannot capture all the necessary information for accurate HD map construction. 
% \end{enumerate}
% The full model (ours), combining SD, DAM, and UAG, achieves the highest performance, demonstrating that leveraging complementary cues from both private and cross-modal experts leads to the best results.

\textbf{Analysis on Expert Group Ablation.} Additionally, we perform an expert group ablation to evaluate the importance of different expert configurations, as shown in Tab.~\ref{tab:expert-group}. We design the following configurations: 
(1) \textbf{Private Experts Only (L+I)}: using only the \textit{private experts from LiDAR and Image}; 
(2) \textbf{Cross-Modal Experts Only (S+Int)}: using only the \textit{cross-modal experts}. 
% The results show that using only private experts (L+I) achieves slightly better performance than the baseline MapTR(+1.3\%), but still underperforms compared to the full model, indicating that relying solely on private experts limits the model’s ability to utilize complementary information. Using only cross-modal experts (S+Int) leads to even weaker performance($-1.9$\% compared to baseline), confirming that cross-modal experts alone cannot capture all the necessary information for accurate HD map construction. The \textbf\textit{{SEF-MAP }}, combining SD, DAM, and UAG, achieves the highest performance, demonstrating that leveraging complementary cues from both private and cross-modal experts leads to the best results.
The results show that using only private experts (L+I) achieves slightly better performance than the baseline MapTR (+1.3\% mAP), with improvements particularly evident in \(AP_{\text{ped}}\) (+1.8\%) and \(AP_{\text{divider}}\) (+1.9\%), suggesting that modality-specific features are valuable for capturing fine-grained semantic and geometric details. 
% However, this configuration still significantly underperforms compared to the full model (-2.9\% mAP), indicating that relying solely on private experts limits the model's ability to utilize complementary cross-modal information and resolve spatial ambiguities that individual modalities cannot handle effectively. 
Using only cross-modal experts (S+Int) leads to even weaker performance (-1.9\% compared to baseline), with notable drops in \(AP_{\text{boundary}}\) (-2.6\%) and \(AP_{\text{ped}}\) (-1.2\%), confirming that cross-modal experts alone cannot capture all the necessary modality-specific cues for accurate HD map construction, such as LiDAR's precise depth information or camera's rich texture details. 
% This demonstrates that shared and interaction features, while important for cross-modal consensus, are insufficient without the foundation of modality-private knowledge.
The \textbf{\textit{SEF-MAP}}, combining SD, DAM, and UAG, achieves the highest performance (+4.2\% mAP), with consistent improvements across all map elements, demonstrating that the synergy between private and cross-modal experts is crucial for leveraging both modality-specific strengths and cross-modal interactions to achieve robust multimodal fusion.

% In Fig.~4, we present qualitative results on a sample scene from nuScenes, showing both LiDAR and camera inputs. We compare the predicted vectorized HD map results of different models, including the baseline MapTR \cite{maptr}, our model with SD only, and the full model (with SD, DAM, and UAG). The baseline model produces significant errors in the map prediction. The joint use of SD and DAM provides the most substantial improvement, allowing the model to effectively fuse information from both modalities and enabling each modality to perform its role. UAG further refines the results, but its impact is more auxiliary, helping to adjust expert contributions rather than directly improving map accuracy. These results confirm that the combined use of SD, DAM, and UAG leads to the best performance by enhancing multi-modal fusion and refining expert contributions.
\subsection{Qualitative Results}

In Fig~\ref{fig:quali}, we show qualitative results on two sample scenes from the nuScenes dataset with both LiDAR and camera inputs. We compare the predicted vectorized HD map results of different models, including the baseline MapTR \cite{maptr}, our model with SD and DAM, and the full model (with SD, DAM, and UAG). The baseline model produces significant errors in map prediction, particularly evident in the red-boxed regions where lane boundaries are misaligned or completely missing. Our progressive model variants show increasingly better performance, with the SD and DAM combination providing substantial improvements in lane detection and road topology preservation. The full SEF-MAP model achieves the best results, demonstrating superior alignment with ground truth and successful reconstruction of complex lane structures and road boundaries that were problematic for the baseline. These visual results confirm that the synergistic combination of SD, DAM, and UAG effectively enhances multi-modal fusion, ultimately producing more reliable and accurate HD map predictions.

\section{CONCLUSION}
% In this work, we propose \textbf{\textit{SEF-MAP}}, a Subspace-Expert Fusion framework for robust multi-modal HD map prediction. \textbf{\textit{SEF-MAP}} decomposes BEV features into LiDAR-private, Image-private, Shared, and Interaction subspaces, each handled by dedicated experts to balance modality-specific information and cross-modal consensus. An uncertainty-aware gating mechanism adaptively regulates expert contributions, while distribution-aware masking with specialization constraints improves robustness under degraded conditions. Overall, \textbf{\textit{SEF-MAP}} offers an effective solution for reliable multi-modal HD map prediction.

In this work, we propose \textbf{\textit{SEF-MAP}}, a Subspace-Expert Fusion framework for robust multi-modal HD map prediction. \textbf{\textit{SEF-MAP}} decomposes BEV features into LiDAR-private, Image-private, Shared, and Interaction subspaces, each handled by dedicated experts to balance modality-specific information and cross-modal consensus. An uncertainty-aware gating mechanism adaptively regulates expert contributions, while distribution-aware masking with specialization constraints improves robustness under degraded conditions. Extensive experiments on nuScenes and Argoverse2 datasets demonstrate the effectiveness of our approach, achieving significant improvements of +4.2\% and +4.8\% mAP respectively over state-of-the-art baselines. The principled design of expert specialization and adaptive fusion not only enhances performance but also provides interpretable insights into modality contributions. Overall, \textbf{\textit{SEF-MAP}} offers an effective solution for reliable multi-modal HD map prediction under diverse and challenging real-world scenarios.
% In this work, we presented SEF-MAP, a Subspace-Expert Fusion framework designed for robust multi-modal HD map prediction. SEF-MAP explicitly decomposes BEV representations into LiDAR-private, Image-private, Shared, and Interaction subspaces, each equipped with dedicated experts. This design enables the model to preserve modality-specific advantages while effectively capturing cross-modal consensus. Furthermore, we introduced an uncertainty-aware gating mechanism that adaptively regulates expert contributions, ensuring reliability across challenging scenarios. To enhance robustness under modality degradation, we proposed distribution-aware masking combined with specialization constraints, encouraging experts to focus on complementary roles. Overall, SEF-MAP provides a principled and effective solution for robust multi-modal fusion, paving the way for more reliable HD map prediction in complex real-world environments.

% \clearpage
\bibliographystyle{IEEEtran}
% \nocite{*}
\bibliography{references}
\end{document}